\documentclass{article}
\usepackage{arxiv}

\usepackage[utf8]{inputenc} % allow utf-8 input
\usepackage[T1]{fontenc}    % use 8-bit T1 fonts
\usepackage{hyperref}       % hyperlinks
\usepackage{url}            % simple URL typesetting
\usepackage{booktabs}       % professional-quality tables
\usepackage{amsfonts}       % blackboard math symbols
\usepackage{nicefrac}       % compact symbols for 1/2, etc.
\usepackage{microtype}      % microtypography
\usepackage{lipsum}		% Can be removed after putting your text content
\usepackage{graphicx}
\usepackage{natbib}
\usepackage{doi}
\usepackage{graphicx}
\usepackage{amsmath}
\usepackage{algorithm}
\usepackage{algpseudocode}
\usepackage{multirow}
\usepackage{graphicx}
\usepackage{subcaption}
\usepackage{listings}

\usepackage{xcolor}

\definecolor{codegreen}{rgb}{0,0.6,0}
\definecolor{codegray}{rgb}{0.5,0.5,0.5}
\definecolor{codepurple}{rgb}{0.58,0,0.82}
\definecolor{backcolour}{rgb}{0.95,0.95,0.92}

\lstdefinestyle{mystyle}{
    backgroundcolor=\color{backcolour},   
    commentstyle=\color{codegreen},
    keywordstyle=\color{magenta},
    numberstyle=\tiny\color{codegray},
    stringstyle=\color{codepurple},
    basicstyle=\ttfamily\footnotesize,
    breakatwhitespace=false,         
    breaklines=true,                 
    captionpos=b,                    
    keepspaces=true,                 
    numbers=left,                    
    numbersep=5pt,                  
    showspaces=false,                
    showstringspaces=false,
    showtabs=false,                  
    tabsize=2
}

\title{\texttt{xai\_evals} : A Framework for Evaluating Post-Hoc Local Explanation Methods}

\author{ 
Pratinav Seth \thanks{Corresponding Author - pratinav.seth@aryaxai.com} , Yashwardhan Rathore , Neeraj Singh, Chintan Chitroda , Vinay Kumar Sankarapu \\
AryaXAI.com, Arya.ai\\
Mumbai, India}

\renewcommand{\headeright}{Technical Report}
\renewcommand{\undertitle}{Technical Report}
\renewcommand{\shorttitle}{\texttt{xai\_evals} : Evaluating Post-Hoc Local Explanation Methods}

\hypersetup{
pdftitle={xai\_evals : A Framework for Evaluating Post-Hoc Local Explanation Methods},
pdfsubject={q-bio.NC, q-bio.QM},
pdfauthor={Pratinav Seth},
pdfkeywords={Explainable AI, Post-hoc Explanation, Model Interpretability, Attribution Methods, Explanation Metrics},
}

\begin{document}
\maketitle

\begin{abstract}
The growing complexity of machine learning and deep learning models has led to an increased reliance on opaque "black box" systems, making it difficult to understand the rationale behind predictions. This lack of transparency is particularly challenging in high-stakes applications where interpretability is as important as accuracy. Post-hoc explanation methods are commonly used to interpret these models, but they are seldom rigorously evaluated, raising concerns about their reliability. The Python package \texttt{xai\_evals} addresses this by providing a comprehensive framework for generating, benchmarking, and evaluating explanation methods across both tabular and image data modalities. It integrates popular techniques like SHAP, LIME, Grad-CAM, Integrated Gradients (IG), and Backtrace, while supporting evaluation metrics such as faithfulness, sensitivity, and robustness. \texttt{xai\_evals} enhances the interpretability of machine learning models, fostering transparency and trust in AI systems. The library is open-sourced at \url{https://pypi.org/project/xai-evals/}.
\end{abstract}

% keywords can be removed
\section{Introduction}
The increasing complexity of machine learning (ML) and deep learning (DL) models has led to their widespread adoption in numerous real-world applications. However, as these models become more powerful, they also become less interpretable. In particular, deep neural networks (DNNs), which have achieved state-of-the-art performance in tasks such as image recognition, natural language processing, and autonomous driving, are often viewed as "black box" models due to their complexity and lack of transparency. 

Interpretability is essential, particularly in high-stakes fields where the consequences of incorrect or non-explainable decisions can be profound. In domains such as healthcare, finance, and law, it is not only crucial that AI systems make accurate predictions but also that these predictions can be understood and justified by human stakeholders. For example, in healthcare, understanding why a model predicts a certain diagnosis can be as important as the prediction itself, influencing clinical decisions and patient outcomes. Similarly, in finance, ensuring that a model's decision-making process is transparent is essential for regulatory compliance and trust.
This growing need for model transparency has led to the development of various explainability techniques aimed at making these "black box" models more understandable. However, despite the availability of these methods, there is a lack of standardized tools for evaluating the quality, robustness, and trustworthiness of the explanations. 

In response to the growing need for explainable machine learning models, we present \texttt{xai\_evals}, a comprehensive Python package designed to facilitate the generation, benchmarking, and evaluation of model explanations. The primary goal of \texttt{xai\_evals} is to provide researchers and practitioners with a unified framework for assessing and comparing various explainability methods. We integrate a range of popular explainability techniques, such as SHAP \cite{Lundberg2017AUA}, LIME \cite{Ribeiro2016WhySI}, Grad-CAM \cite{Selvaraju2016GradCAMVE}, and Integrated Gradients \cite{Sundararajan2017AxiomaticAF}, making it compatible with various model types, including classical machine learning models and deep learning architectures. Additionally, the package includes a robust set of evaluation metrics, including faithfulness, comprehensiveness, and more, that allow for a quantitative assessment of the quality of generated explanations. By providing both explanation generation and evaluation capabilities, \texttt{xai\_evals} aims to foster trust and understanding in deep learning models, enabling their safe and transparent deployment in critical applications.

This paper outlines the functionalities of \texttt{xai\_evals}, presents its core features, and demonstrates how it can be used to generate, evaluate, and benchmark explanations for both tabular and image data models. By providing an easy-to-use tool for explainability, \texttt{xai\_evals} fills a significant gap in the current landscape of model interpretability.

The rest of the paper is structured as follows: Section 2 provides a review of related work in the area of model explainability, explaining the existing techniques and evaluation methods. Section 3 introduces the \texttt{xai\_evals} package in detail, describing its core features, supported explanation methods, and evaluation metrics. Section 4 presents experimental results demonstrating the effectiveness of the package, followed by a discussion of challenges and limitations in Section 5. Finally, Section 6 concludes the paper with a summary of contributions and future work.

\section{Related Work}
\subsection{Overview of Explainability Methods}
Machine learning models, especially deep neural networks, are often regarded as "black boxes" because of the difficulty in understanding how they arrive at specific predictions. To address this issue, explainability methods have been developed to make these models more interpretable. These methods can generally be categorized into global explainability and local explainability.

\subsubsection{Global v/s Local Explainbility}

Global explainability aims to provide a high-level understanding of a model’s overall behavior across all inputs. It focuses on identifying which features are most influential in determining the model's predictions over a large dataset, thus offering insights into how the model makes decisions in general. In contrast, local explainability focuses on explaining the model’s decision-making process for individual instances or predictions. Local explainability is particularly useful for understanding specific decisions made by a model in real-world applications, where a user may want to know why a model made a particular prediction for a specific input. A popular technique within local explainability is post-hoc explainability, where the explanation is generated after the model has been trained without altering its internal structure. These methods are crucial for understanding complex, black-box models like deep neural networks, especially in scenarios where interpretability of individual predictions is necessary.

For instance, local explainability methods are commonly applied in medical diagnostics, where a deep learning model might predict the presence of a disease in a patient based on their medical history and test results. A healthcare professional might want to know why the model predicted that a patient has a particular disease, which can be addressed by local explainability methods. Similarly, in credit scoring models, a financial institution might use local explainability to understand the rationale behind an individual's credit score prediction, helping to provide transparency for both customers and regulators.

\subsubsection{Post Hoc Local Explainibility MEthods}
There are several widely used methods to achieve local explainability, particularly for post-hoc explanations. These methods include:

SHAP (SHapley Additive exPlanations) \cite{Lundberg2017AUA} is a model-agnostic method that assigns a value to each feature, reflecting its contribution to the prediction for a specific instance. SHAP is based on Shapley values from cooperative game theory, ensuring fairness and consistency in the attribution of feature importance. For example, in a loan approval model, SHAP can help explain why a specific feature, such as income, had a significant impact on the decision to approve or deny the loan for a particular applicant. This method can be applied to a variety of models, including tree-based models and deep learning architectures, and is widely used due to its strong theoretical foundation.

LIME (Local Interpretable Model-Agnostic Explanations) \cite{Ribeiro2016WhySI} is another widely used model-agnostic method. LIME approximates the local decision boundary of the model for individual predictions by training an interpretable surrogate model, such as a linear regression or decision tree, on perturbed data. It works by observing how the model behaves on slightly altered versions of the input data and generating an explanation based on this local behavior. For instance, in image classification, LIME can help explain which parts of an image are most influential in determining the model's prediction by perturbing different regions of the image and observing the model's response.

Grad-CAM (Gradient-weighted Class Activation Mapping) \cite{Selvaraju2016GradCAMVE} is a method designed for convolutional neural networks (CNNs) that generates visual explanations by highlighting the regions of an image that are most important in the model's prediction. For example, in a medical imaging task where a CNN is used to diagnose diseases from X-ray images, Grad-CAM can generate a heatmap showing the areas of the X-ray image that contributed most to the prediction of a particular disease, providing intuitive visual explanations to the medical professionals.

Integrated Gradients \cite{Sundararajan2017AxiomaticAF} works by integrating the gradients of the model’s output with respect to the input features, from a baseline input to the actual input. This method provides a smooth and consistent explanation for model predictions, ensuring that each feature’s contribution is computed along a continuous path. For example, in sentiment analysis, Integrated Gradients can be used to explain which words in a sentence were most responsible for a positive or negative sentiment prediction, helping to clarify the rationale behind the model's decision.

Backpropagation-Based Explainability Methods, such as DlBacktrace \cite{sankarapu2024DlBacktracemodelagnosticexplainability}, trace the relevance of each component in a neural network from the output back to the input. This allows for a detailed analysis of how each layer in the network contributes to the final prediction. For example, in image classification, backpropagation-based methods can show how the features learned by the lower layers of the neural network (e.g., edges and textures) contribute to higher-level features (e.g., shapes or objects) that ultimately determine the model’s output. DlBacktrace is particularly useful as it provides insights into the layer-wise contribution to predictions.

These local explainability methods help demystify the decision-making processes of machine learning models, particularly in complex, real-world scenarios. Whether used to explain a healthcare model's diagnosis, a financial institution's credit decision, or a computer vision model's image classification, these techniques provide crucial insights into individual predictions, enhancing the trust and transparency of AI systems.

\subsection{Existing Evaluation Frameworks and Benchmarking Tools for Model Explainability}
Despite significant advancements in Explainable AI (XAI), evaluating model explanations remains a challenge due to the lack of a standardized and comprehensive assessment framework. Existing evaluation methodologies often rely on simplistic interpretability metrics that fail to capture essential aspects such as robustness, generalizability, and human alignment. This has led to inconsistencies in how explanations are assessed across different domains and applications.  

Recent research has highlighted the limitations of current evaluation paradigms. Madsen et al. \cite{madsen2024interpretabilityneedsnewparadigm} argue that most existing approaches prioritize faithfulness while overlooking robustness and usability, leading to incomplete assessments of explanation quality. Wickstrøm et al. \cite{wickstrøm2024flexibilitymanipulationslipperyslope} further emphasize that interpretability metrics are often inconsistent across domains, which complicates their application in real-world scenarios and raises concerns about their susceptibility to manipulation.  

Several benchmarking tools have been introduced to address these issues, each with its own strengths and limitations. The M4 Benchmark \cite{10.5555/3666122.3666203} provides a structured framework for evaluating feature attribution methods, placing significant emphasis on faithfulness. However, it does not explicitly assess robustness against adversarial perturbations or stability across different data distributions, which are crucial for ensuring reliability in high-stakes applications. OpenXAI \cite{agarwal2024openxaitransparentevaluationmodel} offers a flexible evaluation framework, though its reliance on synthetic data generation raises concerns about the generalizability of its findings to real-world settings. Quantus \cite{hedström2023quantusexplainableaitoolkit} incorporates a diverse set of evaluation metrics, covering faithfulness, robustness, and complexity, yet it lacks an explicit mechanism to assess whether the generated explanations align with human intuition. FairX \cite{sikder2024fairxcomprehensivebenchmarkingtool} extends evaluation to fairness and bias considerations but does not provide a comprehensive framework for post-hoc explainability. Similarly, Captum \cite{kokhlikyan2020captumunifiedgenericmodel} and TF-Explain \cite{Meudec2021-le} focus on generating explanations for deep learning models but do not include built-in benchmarking capabilities to assess explanation quality systematically. Inseq \cite{Sarti_2023}, while valuable for sequence generation models, is specialized for NLP tasks and does not generalize well to other domains such as tabular or vision-based data.  

The fragmentation of existing evaluation frameworks highlights the need for a more robust and flexible approach to assessing model explanations. Many existing tools prioritize faithfulness while neglecting complementary factors such as robustness, sensitivity, and usability, leading to an incomplete understanding of explanation quality. Others are designed for specific model architectures or data modalities, making it difficult to conduct cross-domain comparisons. Additionally, many commonly used evaluation metrics do not align well with human judgment, which limits their applicability in decision-critical environments where interpretability is essential.  

To address these challenges, we introduce \texttt{xai\_evals}, a framework that integrates explanation generation and evaluation into a single, standardized package. Unlike existing tools, \texttt{xai\_evals} allows researchers to systematically assess explanation quality across multiple methods, models, and datasets using a broad range of evaluation metrics. 

\section{Package Overview}
\begin{figure}
    \centering
    \includegraphics[width=0.99\linewidth]{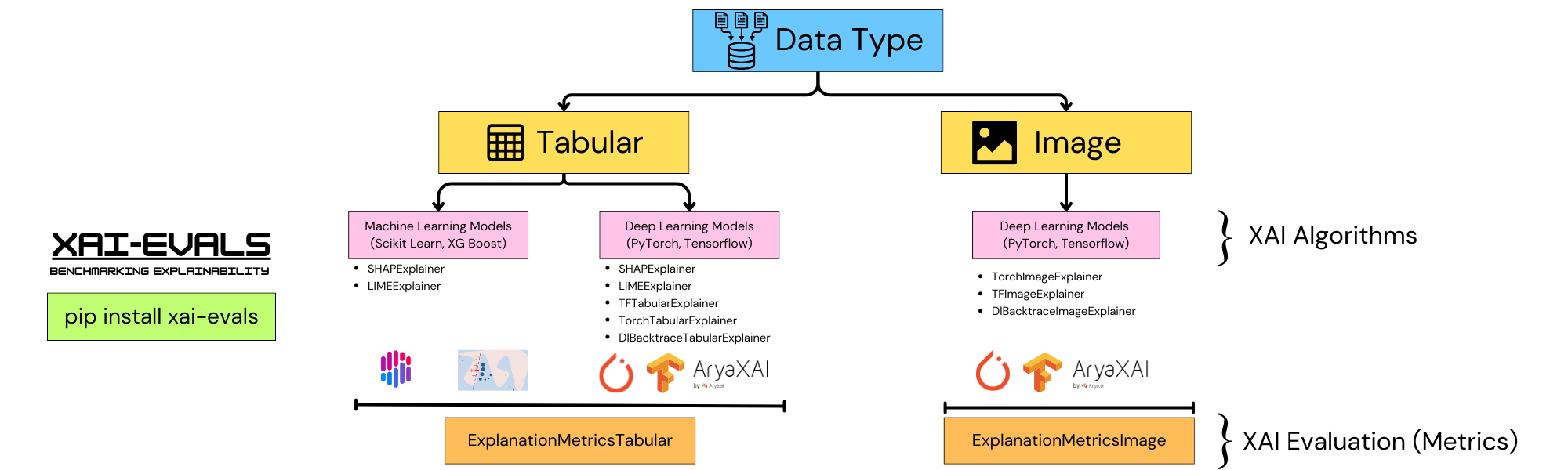}
    \caption{\texttt{xai\_evals} Library Overview.}
    \label{fig:evals_lib}
\end{figure}
The \texttt{xai\_evals} package provides a comprehensive suite of functionalities to facilitate model interpretability and explainability. The main features of the package include:

\begin{itemize}
    \item \textbf{Compatibility with Various Models:} The package supports both classical machine learning models, such as those in scikit-learn (e.g., RandomForest, LogisticRegression, etc.), as well as deep learning models built using frameworks like PyTorch and TensorFlow.
    \item \textbf{Model Explanation Generation:} The package integrates several popular explainability methods, including SHAP, LIME, Grad-CAM, Integrated Gradients, and Backtrace, to generate local and global explanations for a wide range of machine learning models.
    \item \textbf{Model Explanation Evaluation:} \texttt{xai\_evals} offers a set of robust evaluation metrics, such as faithfulness, sensitivity, comprehensiveness, and robustness, to quantitatively assess the quality of generated explanations.
    \item \textbf{Benchmarking Explanations:} The package allows for benchmarking explanations generated by different methods, facilitating a comparison of explanation quality across different models and datasets.
    
\end{itemize}

\subsection{Model and Data Type Support}
\begin{table}[pb]
\centering
\small
\caption{Comparison of Illustration and Metric Classes in \texttt{xai\_evals}.}
\begin{tabular}{|c|c|c|}
\hline
\textbf{Category}          & \textbf{Illustration (Explainability)}                                                                                                  & \textbf{Evaluation (Metrics)}  \\ \hline
\textbf{Function}          & Generates Explainibility (Feature Attributions)                                                                                         & Quantifies explanation quality \\ \hline
\textbf{Output}            & Explanations                                                                                                                            & Numeric Metric Values          \\ \hline
\textbf{Example (Tabular)} & \begin{tabular}[c]{@{}c@{}}\texttt{SHAPExplainer,DlBacktraceTabularExplainer},\\ \texttt{LIMEExplainer,TorchImageExplainer,TFImageExplainer}\end{tabular} & \texttt{ExplanationMetricsTabular}      \\ \hline
\textbf{Example (Image)}   & \begin{tabular}[c]{@{}c@{}}\texttt{TorchImageExplainer,TFImageExplainer},\\ \texttt{DlBacktraceImageExplainer}\end{tabular}                               & \texttt{ExplanationMetricsImage}        \\ \hline
\end{tabular}
\end{table}

The \texttt{xai\_evals} framework is structured to support explainability for \textbf{tabular} and \textbf{image} data across \textbf{machine learning (ML)} and \textbf{deep learning (DL)} models.
\subsubsection{Tabular Data}
\begin{itemize}
    \item \textbf{Machine Learning Models (Scikit-Learn, XGBoost)} : \texttt{SHAPExplainer}, \texttt{LIMEExplainer}.
    \item \textbf{Deep Learning Models}: \texttt{SHAPExplainer}, \texttt{LIMEExplainer}, \texttt{TFTabularExplainer}, \texttt{TorchTabularExplainer}, \texttt{DlBacktraceTabularExplainer}.
    \item \textbf{Evaluation:} \texttt{ExplanationMetricsTabular}.
\end{itemize}

\subsubsection{Image Data}
\begin{itemize}
    \item \textbf{Deep Learning Models}: \texttt{TorchImageExplainer}, \texttt{TFImageExplainer}, \texttt{DlBacktraceImageExplainer}.
    \item \textbf{Evaluation:} \texttt{ExplanationMetricsImage}.
\end{itemize}
\subsection{Explanation vs. Evaluation Classes}
The \texttt{xai\_evals} framework consists of two key components: \textbf{illustration classes} for generating explanations and \textbf{metric classes} for evaluating their quality.

\subsection{Installation}

To install the \texttt{xai\_evals} package, you can either clone the repository from GitHub or install it directly via pip (once the package is published). Below are the instructions for both the methods:

\subsubsection{Install via Pip (After Package is Published)}

You can install it directly from pip using the following command:

\begin{lstlisting}
pip install xai_evals
\end{lstlisting}

\subsection{Explanation Methods Supported}

The \texttt{xai\_evals} package integrates a wide variety of explanation methods, which can be used to generate attributions for models. These methods can be categorized into those for tabular data and image data, as well as methods tailored for deep learning versus classical machine learning models.

\subsubsection{Tabular Data Explanations}

The package provides several explanation methods for tabular data. For Machine Learning models, it supports SHAP and LIME-based explainers, tailored to the specific model type. For Deep Learning models, both LIME and SHAP are supported, alongside a variety of other methods such as integrated gradients, deep LIFT, gradient SHAP, saliency, input $X$ gradient, guided backprop, SHAP Kernel, SHAP Deep, and LIME for PyTorch models. Additionally, the package includes support for DlBacktrace, which works with both PyTorch and TensorFlow models.
\subsubsection{Image Data Explanations}
The package offers various explanation methods for image data. For Torch, it supports methods such as Grad-CAM, integrated gradients, saliency, deep LIFT, gradient SHAP, guided backprop, occlusion, layer Grad-CAM, and feature ablation. For TensorFlow, it provides Vanilla Gradients, Grad-CAM, Gradients Input, Integrated Gradients, Occlusion Sensitivity, and SmoothGrad. Additionally, for Backtrace, the package supports the methods default, contrast-positive, and contrast-negative.

\subsection{Metrics for Evaluation}
\subsubsection{Tabular}
To assess the quality of generated explanations, the \texttt{xai\_evals} library supports several evaluation metrics. These metrics help quantify how well the explanations align with the model’s predictions and the contributions of individual features. Below, we describe each metric along with its mathematical formulation and interpretation.

\textbf{Faithfulness} measures how well the attribution aligns with the changes in model output when features are perturbed. The metric assesses whether the attribution reflects the changes in the model’s output when a given feature is modified. A higher faithfulness score indicates that the attribution values align well with the changes in model output caused by feature perturbations, suggesting that the explanation is an accurate reflection of the model’s behavior. Low faithfulness implies that the explanation may not be faithful to how the model actually reacts to input changes. Mathematically, it can be defined as:

\[
\text{Faithfulness} = \frac{1}{n} \sum_{i=1}^{n} \left| f(x) - f(x_i') \right| \cdot \left| a_i \right|
\]

Where: \( f(x) \) is the original model output for the input \( x \), \( f(x_i') \) is the model output after perturbing the \(i\)-th feature in \( x \), \( a_i \) is the attribution for the \(i\)-th feature, \( n \) is the total number of features.

\textbf{Sensitivity} measures the robustness of the attributions to small perturbations in the input features. This metric evaluates how much the attributions change when small random noise is added to the input features. Low sensitivity scores suggest that the attributions are stable and not affected by small noise in the input data, which is generally desirable for robustness. High sensitivity indicates that the explanation is unstable, and small changes in the input can lead to large variations in attributions, which may imply that the model’s explanations are not robust. The formula is:

\[
\text{Sensitivity} = \frac{1}{n} \sum_{i=1}^{n} \left| a_i - a_i' \right|
\]

Where \( a_i \) is the attribution score for the \(i\)-th feature on the original input \( x \), \( a_i' \) is the attribution score after a small perturbation is applied to the \(i\)-th feature in \( x \).

\textbf{Comprehensiveness}  measures how much the model’s output decreases when the most important features, as identified by the explanation, are removed. A high comprehensiveness score indicates that the removal of the most important features significantly affects the model's prediction, suggesting that these features are crucial for the model’s decision-making process. Low comprehensiveness suggests that the explanation is incomplete and removing the important features does not substantially impact the output. The formula is:

\[
\text{Comprehensiveness} = \frac{1}{n} \sum_{i=1}^{n} \left( f(x) - f(x_{\text{mask}}) \right) \cdot \mathbb{1}_{\{a_i \in S_k\}}
\]

Where \( f(x) \) is the model’s original output, \( f(x_{\text{mask}}) \) is the model output after masking the top \(k\) features based on their attributions, \( \mathbb{1}_{\{a_i \in S_k\}} \) is an indicator function that equals 1 if the feature \( a_i \) belongs to the set \( S_k \) of the most important features.

\textbf{Sufficiency} measures whether the most important features alone are enough to explain the model’s output. A high sufficiency score indicates that the most important features alone can approximate the original prediction, suggesting that these features are sufficient to explain the model’s decision. If the sufficiency score is low, it implies that additional features beyond the top \(k\) are needed for an adequate explanation. The metric is given by:

\[
\text{Sufficiency} = \frac{1}{n} \sum_{i=1}^{n} \left( f(x) - f(x_{\text{focused}}) \right) \cdot \mathbb{1}_{\{a_i \in S_k\}}
\]

Where \( f(x) \) is the model’s original output, \( f(x_{\text{focused}}) \) is the model output when only the top \(k\) important features are retained and all others are set to zero.

\textbf{Monotonicity} checks if the attributions are consistent with the direction of change in the model output. That is, attributions should increase or decrease as the model’s prediction increases or decreases. A high monotonicity score suggests that the attributions change in a consistent direction with the model’s output, which implies a coherent and logically consistent explanation. A low score may indicate that the explanation does not align with the model’s underlying logic. The formula is:

\[
\text{Monotonicity} = \frac{1}{n-1} \sum_{i=1}^{n-1} \mathbb{1}_{\{ \text{sign}(a_i) = \text{sign}(a_{i+1}) \}}
\]

Where \( a_i \) is the attribution score for the \(i\)-th feature, The \(\text{sign}(a_i)\) is the sign of the attribution (positive or negative).

\textbf{Complexity} measures the sparsity of the explanation by counting the number of features with non-zero attribution values. A low complexity score suggests a sparse and interpretable explanation where only a few features are important for the model’s prediction. A high complexity score suggests that the model relies on many features, which may make the explanation more difficult to interpret. It is given by:

\[
\text{Complexity} = \frac{1}{n} \sum_{i=1}^{n} \mathbb{1}_{\{ a_i \neq 0 \}}
\]

Where \( a_i \) is the attribution score for the \(i\)-th feature.

\textbf{Sparseness} measures the minimalism of the explanation by calculating the proportion of features that have zero attribution values. A high sparseness score suggests that the explanation is minimal, with most features receiving zero attribution, indicating that only a few features are considered important. A low sparseness score indicates that many features are being attributed, leading to a more complex and potentially harder-to-interpret explanation. It is given by:

\[
\text{Sparseness} = 1 - \frac{1}{n} \sum_{i=1}^{n} \mathbb{1}_{\{ a_i \neq 0 \}}
\]

Where \( a_i \) is the attribution score for the \(i\)-th feature.

\subsubsection{Image}

To assess the quality of generated explanations for image models, the \texttt{xai\_evals} library supports several evaluation metrics. These metrics help quantify how well the explanations align with the model's predictions and the contributions of individual image features. Below, we describe each metric along with its mathematical formulation and interpretation.

\textbf{Faithfulness Correlation }\cite{Bhatt2020EvaluatingAA} measures how well the attribution aligns with the changes in model output when pixels or regions of the image are perturbed. A high Faithfulness Correlation score indicates that the explanation is faithful to the model’s behavior, meaning that the attributions reflect the actual changes in model output due to perturbations. A low score suggests that the attribution method may not accurately represent the model’s sensitivity to changes in the image. The metric is defined as:

\[
\text{Faithfulness Correlation} = \frac{\sum_{i=1}^{n} \left| f(x) - f(x_i') \right| \cdot \left| a_i \right|}{\sum_{i=1}^{n} \left| a_i \right|}
\]

Where: 
\( f(x) \) is the model’s original output for the input image \( x \), 
\( f(x_i') \) is the model output after perturbing the \(i\)-th pixel or region, 
\( a_i \) is the attribution value for the \(i\)-th pixel or region, 
and \( n \) is the total number of pixels or regions in the image.

\textbf{Max Sensitivity} \cite{Yeh2019OnT} evaluates how much the model’s output changes when specific pixels or regions of the image are perturbed. It provides a measure of how sensitive the model is to changes in individual features of the image.  A higher Max Sensitivity score indicates that the model's output is highly sensitive to changes in specific image regions, suggesting that those regions play a key role in the model’s decision. A lower score indicates less sensitivity, meaning that perturbing specific pixels has less impact on the model's prediction. The formula is:

\[
\text{Max Sensitivity} = \max_{i=1}^{n} \left| f(x) - f(x_i') \right|
\]

Where:
\( f(x) \) is the model’s original output, 
and \( f(x_i') \) is the model output after perturbing the \(i\)-th pixel or region.

\textbf{MPRT (Mean Pixelwise Robustness)} \cite{Adebayo2018SanityCF} measures the robustness of the model’s predictions against perturbations of individual pixels. This metric computes the average change in the model output when each pixel is individually perturbed. A lower MPRT score indicates that the model is robust to perturbations, meaning the explanation is stable and not influenced by small changes. A higher MPRT score suggests that the model is sensitive to pixel perturbations, which might indicate that the explanation is less reliable or that the model is overfitting to specific features. The formula is:

\[
\text{MPRT} = \frac{1}{n} \sum_{i=1}^{n} \left| f(x) - f(x_i') \right|
\]

Where:
\( f(x) \) is the model’s original output, 
and \( f(x_i') \) is the model output after perturbing the \(i\)-th pixel.

\textbf{Smooth MPRT} \cite{Hedstrm2024SanityCR} is a variation of MPRT that smoothens the impact of perturbations by considering both the perturbation’s effect on the model’s output and the magnitude of the attributions. Smooth MPRT helps to balance the effect of perturbations with the importance of each pixel (as indicated by its attribution). A higher Smooth MPRT score means the model’s output is highly sensitive to pixels with high attribution values. A lower score suggests that the model’s output is more stable and that the attributions are less sensitive to perturbations. It is defined as:

\[
\text{Smooth MPRT} = \frac{1}{n} \sum_{i=1}^{n} \frac{\left| f(x) - f(x_i') \right|}{1 + \left| a_i \right|}
\]

Where:
\( f(x) \) is the model's original output, 
\( f(x_i') \) is the model output after perturbing the \(i\)-th pixel, 
and \( a_i \) is the attribution value for the \(i\)-th pixel.

\textbf{Avg Sensitivity} \cite{Yeh2019OnT} measures the average sensitivity of the model’s output to pixel perturbations across all pixels in the image. This metric gives an overall sense of how much the model’s output changes when any pixel or region is perturbed. A lower Avg Sensitivity score indicates that perturbing pixels does not significantly affect the model’s output, suggesting that the explanation is stable. A higher score means that the model’s prediction is highly sensitive to changes in the image, which might indicate that the model is over-relying on certain features. The formula is:

\[
\text{Avg Sensitivity} = \frac{1}{n} \sum_{i=1}^{n} \left| f(x) - f(x_i') \right|
\]

Where:
\( f(x) \) is the model’s original output, 
and \( f(x_i') \) is the model output after perturbing the \(i\)-th pixel.

\textbf{Faithfulness Estimate} \cite{AlvarezMelis2018TowardsRI} evaluates the extent to which the attributions reflect the model’s behavior when perturbations are made to the image. It uses a baseline image (e.g., a black image) for comparison to estimate how well the explanation matches the model’s output changes. A higher Faithfulness Estimate score indicates that the attributions align well with the model's behavior when pixels are perturbed. A lower score suggests that the explanation does not faithfully represent the model's decision-making process. The formula is:

\[
\text{Faithfulness Estimate} = \frac{1}{n} \sum_{i=1}^{n} \left| f(x) - f(x_i') \right| \cdot \left| a_i \right|
\]

Where:
\( f(x) \) is the model’s output for the original image, 
\( f(x_i') \) is the model output after perturbing the \(i\)-th pixel, 
and \( a_i \) is the attribution value for the \(i\)-th pixel.

\subsection{Usage Illustrations}
After installation, you can begin using \texttt{xai\_evals} to generate and evaluate explanations for your machine learning models. The library supports a variety of explainers, such as SHAP, LIME, and Grad-CAM, for different model types and data formats, enabling both illustrative and quantitative comparisons. Below are examples demonstrating how to use the library for both tabular and image data.

\subsubsection{Tabular Data Example:}
Illustration on how to use the SHAP explainer for illustrating attribution for a tabular dataset using a Random Forest Classifier:
\begin{lstlisting}
# Load dataset and train a model
data = load_iris()
X = pd.DataFrame(data.data, columns=data.feature_names)
y = data.target
model = RandomForestClassifier()
model.fit(X, y)

# Initialize SHAP explainer
shap_explainer = SHAPExplainer(model=model, features=X.columns, task="multiclass-classification", X_train=X)

# Explain a specific instance (e.g., the first instance in the test set)
shap_attributions = shap_explainer.explain(X, instance_idx=0)
print(shap_attributions)
\end{lstlisting}
\begin{table}[H]
\centering
\begin{tabular}{cccc}
\cline{1-4}
idx & Feature             & Value & Attribution    \\ \hline
2   & petal\_length\_(cm) & 1.4   & \textbf{0.356} \\
3   & petal\_width\_(cm)  & 0.2   & \textbf{0.309} \\
0   & sepal\_length\_(cm) & 5.1   & 0.022          \\
1   & sepal\_width\_(cm)  & 3.5   & 0.003          \\ \hline
\end{tabular}
\vskip 0.1in
\caption{Illustration of Attribution of Each Feature using SHAP for a Random Forest Model trained on IRIS Dataset whose prediction and label are both 0}
\end{table}

\subsubsection{Image Data Example:}
Illustration on how to use TorchImageExplainer for Generating GradCAM explaination for a Pytorch CNN Model (ResNet18) over a resized Imagenette Dataset Sample.
\begin{lstlisting}
transform = transforms.Compose([transforms.Resize((224, 224)),
                    transforms.ToTensor(),transforms.Normalize(
                    mean=[0.485, 0.456, 0.406], std=[0.229, 0.224, 0.225]),])
                    
imagenette_dataset = datasets.ImageFolder(root=os.path.join(dataset_path, "val"), transform=transform)

model = models.resnet18(weights=models.ResNet18_Weights.IMAGENET1K_V1)
model.eval()
explainer = TorchImageExplainer(model)

input_tensor, _ = imagenette_dataset[9]
attribution_map = explainer.explain(input_tensor, method="layer_gradcam",                                               task="classification")
if isinstance(attribution_map, torch.Tensor):
    attribution_map = attribution_map.squeeze().detach().numpy()
plt.imshow(input_tensor.permute(1, 2, 0).numpy())
plt.imshow(attribution_map, cmap="jet", alpha=0.5)
plt.title("Grad-CAM Attribution Map")
plt.colorbar()
plt.show()
\end{lstlisting}

\begin{figure}[H]
    \centering
    \includegraphics[width=0.3\linewidth]{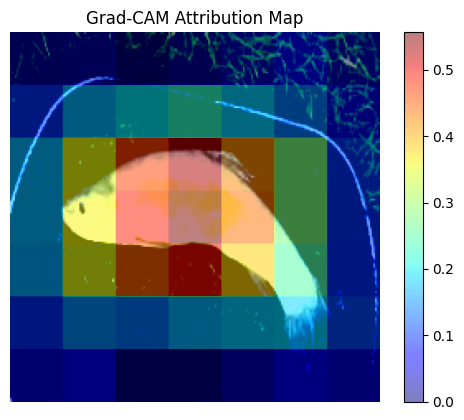}
    \caption{Illustration of Overlay GradCAM attribution map over Image Sample. The attribution map highlights the important regions in the image that contributed to the model's prediction.}
    \label{fig:gradcam_image}
\end{figure}

\subsubsection{Evaluating Explanations:}
\texttt{xai\_evals} allows you to evaluate their quality using a variety of metrics by using ExplanationMetricsTabular and ExplanationMetricsImage. Here’s an example of how to calculate evaluation metrics using ExplanationMetricsTabular for SHAP for a Random Forest Classifier over IRIS Dataset.
\begin{lstlisting}
# Load dataset and train a model
data = load_iris()
X = pd.DataFrame(data.data, columns=data.feature_names)
y = data.target
model = RandomForestClassifier()
model.fit(X, y)

# Initialize ExplanationMetrics with SHAP explainer
explanation_metrics = ExplanationMetricsTabular(
    model=model,
    explainer_name="shap",
    X_train=X,
    X_test=X,
    y_test=y,
    features=X.columns,
    task="binary-classification"
)

# Calculate metrics
metrics_df = explanation_metrics.calculate_metrics()
print(metrics_df)
\end{lstlisting}

\begin{table}[H]
\centering
\label{tab:illustration_iris_shap}
\begin{tabular}{cccccccc}
\hline
faithfulness & infidelity & sensitivity & comprehensiveness & sufficiency & monotonicity & complexity & sparseness \\ \hline
-0.568218    & 0.000048   & 0.00328     & -0.0124           & 0.0         & 0.973333     & 4.0        & 0.0        \\ \hline
\end{tabular}
\vskip 0.1in
\caption{Illustration of Output Metrics Value using ExplanationMetricsTabular over a Random Forest Classifier trained on IRIS Dataset using SHAP Attribution .}
\end{table}

\section{The Importance of Explainable AI (XAI) and Regulatory Compliance}

Explainable AI (XAI) is becoming increasingly important as machine learning models are used in critical fields such as healthcare, finance, and autonomous systems. In these areas, decisions made by AI need to be interpretable to ensure trust, transparency, and accountability. Understanding how a model arrives at its conclusions helps businesses, regulators, and end-users assess its fairness and reliability.
The \texttt{xai\_evals} package provides a way to analyze model decisions by generating explanations that clarify how inputs influence predictions. This is particularly relevant for regulatory compliance, such as the European Union’s \textbf{General Data Protection Regulation (GDPR)} and the forthcoming \textbf{EU AI Act}, which require AI systems to be transparent and accountable. GDPR grants individuals the \textit{"right to explanation"} when automated decisions affect them, making interpretability a legal necessity in many applications.
By using \texttt{xai\_evals}, researchers and practitioners can verify that their models make decisions based on meaningful and justifiable features. This helps organizations not only meet legal requirements but also build trust with stakeholders and ensure AI systems are used responsibly in regulated industries.

\section{Conclusion}
We introduced \texttt{xai\_evals}, a Python package designed to generate, benchmark, and evaluate various model explainability methods. The package aims to bridge the gap between model accuracy and interpretability by providing a comprehensive framework for both machine learning and deep learning models. By supporting popular explainability methods like SHAP, LIME, Grad-CAM, Integrated Gradients, DlBacktrace and more . \texttt{xai\_evals} enables researchers and practitioners to gain deeper insights into their models' decision-making processes. The library also provides a variety of evaluation metrics to assess the quality and reliability of these explanations.

Through our example applications, we demonstrated how \texttt{xai\_evals} can be used to explain and evaluate both tabular and image-based models. The results of our benchmarking experiments showed that \texttt{xai\_evals} is effective in comparing the quality of different explanation methods and provides valuable insights into model behavior.

\section{Future Directions}

While \texttt{xai\_evals} provides a strong foundation for model explainability, there are several key areas for future improvement. Our goal is to expand its capabilities to support a wider range of models, integrate new explanation methods, enhance performance, and improve overall usability.

\subsection{Extending Model Support}

To make \texttt{xai\_evals} more versatile, we plan to extend its compatibility to a broader range of  models, including:

\textbf{Natural Language Processing Models: }
There is increasing reliance on deep learning models for NLP tasks, future updates will introduce explanation techniques for text-based models.

\textbf{Hugging Face Transformers and Autoregressive Models}
Transformer-based architectures, including those provided by Hugging Face (e.g., BERT, GPT, T5, Llama), are widely used across NLP and beyond. Future iterations will integrate explainability methods for these models.

\textbf{Graph Neural Networks (GNNs)}
GNNs are increasingly used in fields like drug discovery, fraud detection, and recommendation systems. 

\subsection{Expanding Explanation Methods \& Enhancing Evaluation Metrics}

While \texttt{xai\_evals} currently supports several widely used explanation techniques, future updates will introduce additional methods to improve interpretability across diverse applications.
To provide more comprehensive assessments of model interpretability, we plan to refine and expand the evaluation metrics .

\subsection{Optimizing Performance, Scalability and Stability Enhancements}

With the increasing complexity of deep learning models, maintaining computational efficiency in explainability methods is essential. Future enhancements to \texttt{xai\_evals} will include GPU acceleration, parallel processing, and memory optimization for large-scale datasets, along with real-time explainability and distributed processing to ensure smooth deployment in production environments.

Ensuring the reliability of \texttt{xai\_evals} is a key focus. Ongoing improvements will include Addressing reported bugs and inconsistencies in the current version. Optimizing code efficiency and reducing memory overhead. Expanding test coverage to improve robustness across different model architectures.

\section*{Acknowledgments}

We extend our sincere gratitude to the creators and maintainers of various open-source libraries that have been instrumental in the development of \texttt{xai\_evals}. In particular, we acknowledge the following contributions:

\begin{itemize}
    \item \textbf{Quantus} \cite{hedström2023quantusexplainableaitoolkit}provides a comprehensive set of metrics for evaluating model explanations in the image modality. In our current version, we utilize it for metric computation in image-based models. While we plan to remove this dependency in future iterations, we deeply appreciate its contributions and support.
    
    \item \textbf{Captum}\cite{kokhlikyan2020captumunifiedgenericmodel}, an open-source library for model interpretability developed by Facebook, offers several attribution methods, including Integrated Gradients, Saliency, DeepLift, and Grad-CAM. We are grateful for these techniques, which we have incorporated into our library to enhance the interpretability of PyTorch models.
    
    \item \textbf{tf-explain} \cite{Meudec2021-le} provides key explanation techniques for TensorFlow/Keras models, such as Grad-CAM and Occlusion Sensitivity. We appreciate the authors of \texttt{tf-explain} for making these tools available, as they have been valuable in explaining TensorFlow-based models.
    
    \item We also acknowledge the developers of \textbf{LIME (Local Interpretable Model-Agnostic Explanations)} \cite{Ribeiro2016WhySI} and \textbf{SHAP (SHapley Additive exPlanations)} \cite{Lundberg2017AUA} for providing foundational baseline methods that are widely used across various domains for model explainability.
\end{itemize}

The open-source nature of these tools has enabled us to integrate them seamlessly into our library, enhancing the explainability of machine learning models.

Additionally, we extend our gratitude to the broader machine learning and AI community for their invaluable discussions, suggestions, and contributions, which have helped shape the direction of \texttt{xai\_evals}. We remain committed to continuously improving the library by expanding its capabilities and making it a more effective tool for evaluating model interpretability.

\bibliographystyle{unsrtnat}
\bibliography{references}

\end{document}